\documentclass[10pt,twocolumn,letterpaper]{article}

\usepackage{cvpr}
\usepackage{times}
\usepackage{epsfig}
\usepackage{graphicx}
\usepackage{amsmath}
\usepackage{amssymb}

\usepackage[bookmarks=false,colorlinks=true,linkcolor=black,citecolor=black,filecolor=black,urlcolor=black]{hyperref}
\usepackage[british,UKenglish,USenglish,english,american]{babel}

\cvprfinalcopy 

\ifcvprfinal\fi
\begin{document}

\title{Instance-aware Semantic Segmentation via Multi-task Network Cascades}

\author{Jifeng Dai \qquad\qquad Kaiming He \qquad\qquad Jian Sun \vspace{8pt}\\
Microsoft Research\\
{\tt\small \{jifdai,kahe,jiansun\}@microsoft.com}
}

\maketitle

\begin{abstract}
\vspace{-1em}
Semantic segmentation research has recently witnessed rapid progress, but many leading methods are unable to identify object instances. In this paper, we present Multi-task Network Cascades for instance-aware semantic segmentation. Our model consists of three networks, respectively differentiating instances, estimating masks, and categorizing objects. These networks form a cascaded structure, and are designed to share their convolutional features. We develop an algorithm for the nontrivial end-to-end training of this causal, cascaded structure. Our solution is a clean, single-step training framework and can be generalized to cascades that have more stages.
We demonstrate state-of-the-art instance-aware semantic segmentation accuracy on PASCAL VOC. Meanwhile, our method takes only 360ms testing an image using VGG-16, which is two orders of magnitude faster than previous systems for this challenging problem. As a by product, our method also achieves compelling object detection results which surpass the competitive Fast/Faster R-CNN systems.

The method described in this paper is the foundation of our submissions to the MS COCO 2015 segmentation competition, where we won the 1st place.
\end{abstract}




\section{Introduction}

Since the development of fully convolutional networks (FCNs) \cite{Long2015}, the accuracy of semantic segmentation has been improved rapidly \cite{Chen2015,Papandreou2015,Dai2015a,Zheng2015} thanks to deeply learned features \cite{Krizhevsky2012,Simonyan2015}, large-scale annotations \cite{Lin2014}, and advanced reasoning over graphical models \cite{Chen2015,Zheng2015}. Nevertheless, FCNs \cite{Long2015} and improvements \cite{Chen2015,Papandreou2015,Dai2015a,Zheng2015} are designed to predict a category label for each pixel, \emph{but are unaware of individual object instances}.
Accurate and fast instance-aware semantic segmentation is still a challenging problem.
To encourage the research on this problem, the recently established COCO \cite{Lin2014} dataset and competition only accept instance-aware semantic segmentation results.

\begin{figure}[t]
\centering
\includegraphics[width=0.9\linewidth]{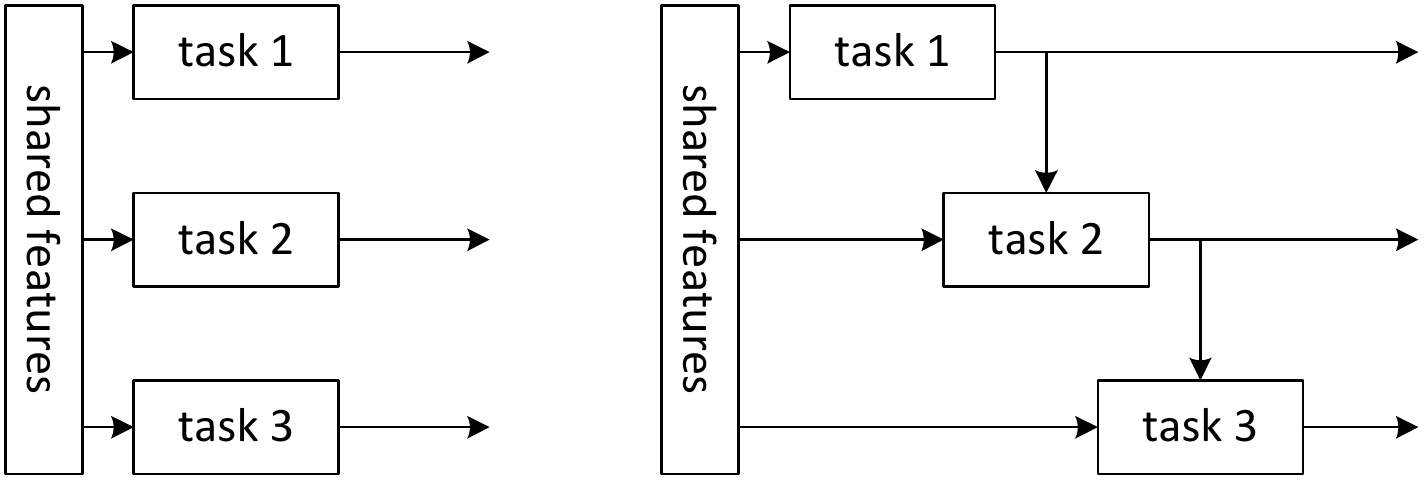}
\caption{Illustrations of common multi-task learning (left) and our multi-task cascade (right).}
\label{fig:illustration}
\vspace{-1em}
\end{figure}

There have been a few methods \cite{Girshick2014,Hariharan2014,Dai2015,Hariharan2015} addressing instance-aware semantic segmentation using convolutional neural networks (CNNs) \cite{LeCun1989,Krizhevsky2012}.
These methods all require mask proposal methods \cite{Uijlings2013,Carreira2012a,Arbelaez2014} that are slow at inference time. In addition, these mask proposal methods take no advantage of deeply learned features or large-scale training data, and may become a bottleneck for segmentation accuracy.

In this work, we address instance-aware semantic segmentation solely based on CNNs, without using external modules (\eg, \cite{Arbelaez2014}).
We observe that the instance-aware semantic segmentation task can be decomposed into three different and related sub-tasks. 1) \emph{Differentiating instances}. In this sub-task, the instances can be represented by bounding boxes that are class-agnostic.
2) \emph{Estimating masks}. In this sub-task, a pixel-level mask is predicted for each instance. 3) \emph{Categorizing objects}. In this sub-task, the category-wise label is predicted for each mask-level instance. We expect that each sub-task is simpler than the original instance segmentation task, and is more easily addressed by convolutional networks.

\begin{figure*}[t]
\centering
\includegraphics[width=0.85\linewidth]{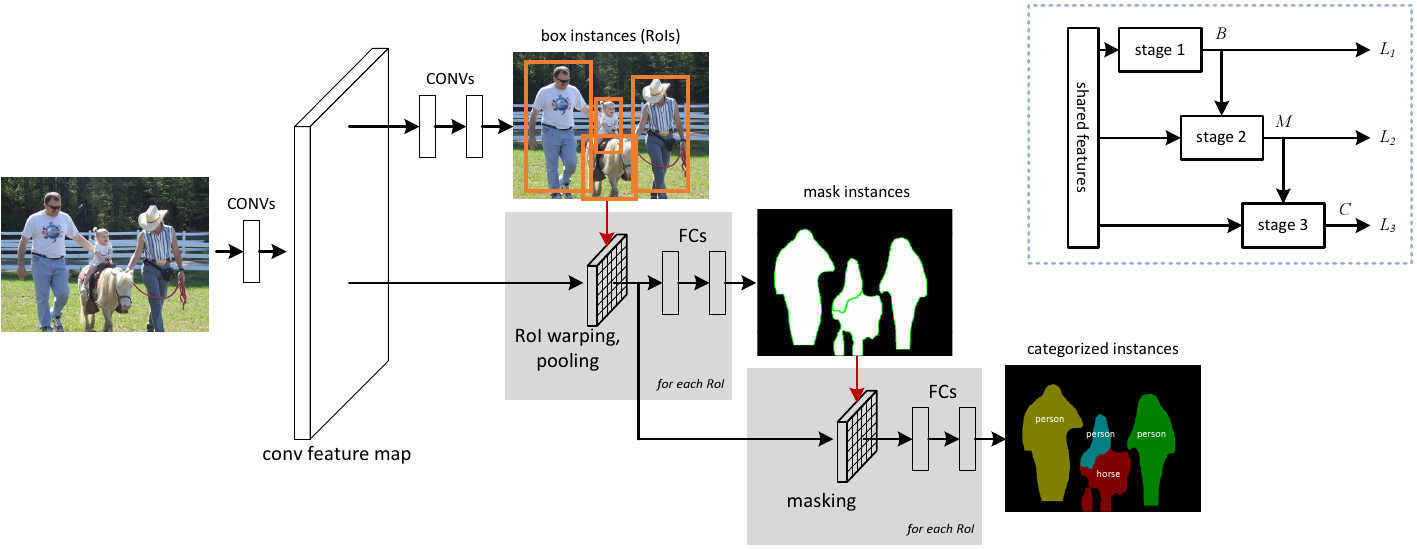}
\caption{Multi-task Network Cascades for instance-aware semantic segmentation. At the top right corner is a simplified illustration.}
\label{fig:overview}
\end{figure*}

Driven by this decomposition, we propose \emph{Multi-task Network Cascades} (MNCs) for accurate and fast instance-aware semantic segmentation. Our network cascades have three stages, each of which addresses one sub-task.
The three stages share their features, as in traditional multi-task learning \cite{Caruana1997}. Feature sharing greatly reduces the test-time computation, and may also improve feature learning thanks to the underlying commonality among the tasks. But \emph{unlike} many multi-task learning applications, in our method a later stage depends on the outputs of an earlier stage, forming a \emph{causal} cascade (see Fig.~\ref{fig:illustration}). So we call our structures ``multi-task cascades''.

Training a multi-task cascade is nontrivial because of the causal relations among the multiple outputs. For example, our mask estimating layer takes convolutional features and predicted box instances as inputs, both of which are outputs of other layers. According to the chain rule of backpropagation \cite{LeCun1989}, the gradients involve those with respect to the convolution responses and also those with respect to the spatial coordinates of predicted boxes. To achieve theoretically valid backpropagation, we develop a layer that is differentiable with respect to the spatial coordinates, so the gradient terms can be computed.

Our cascade model can thus be trained end-to-end via a clean, single-step framework. This single-step training algorithm naturally produces convolutional features that are shared among the three sub-tasks, which are beneficial to both accuracy and speed. Meanwhile, under this training framework, our cascade model can be extended to more stages, leading to improvements on accuracy.

We comprehensively evaluate our method on the PASCAL VOC dataset. Our method results in 63.5\% mean Average Precision (mAP$^r$), about 3.0\% higher than the previous best results \cite{Hariharan2015,Dai2015} using the same VGG network \cite{Simonyan2015}. Remarkably, this result is obtained at a test-time speed of 360ms per image, which is two orders of magnitudes faster than previous systems \cite{Hariharan2015,Dai2015}.

Thanks to the end-to-end training and the independence of external modules, the three sub-tasks and the entire system easily benefit from stronger features learned by deeper models. We demonstrate excellent accuracy on the challenging MS COCO segmentation dataset using an extremely deep 101-layer residual net (ResNet-101) \cite{He2015a}, and also report our \emph{1st-place result in the COCO segmentation track} in ILSVRC \& COCO 2015 competitions.

\section{Related Work}

\emph{Object detection} methods \cite{Girshick2014,He2014,Girshick2015,Ren2015} involve predicting object bounding boxes and categories. The work of R-CNN \cite{Girshick2014} adopts region proposal methods (\eg, \cite{Uijlings2013,Zitnick2014}) for producing multiple instance proposals, which are used for CNN-based classification. In SPPnet \cite{He2014} and Fast R-CNN \cite{Girshick2015}, the convolutional layers of CNNs are shared on the entire image for fast computation. Faster R-CNN \cite{Ren2015} exploits the shared convolutional features to extract region proposals used by the detector. Sharing convolutional features leads to substantially faster speed for object detection systems \cite{He2014,Girshick2015,Ren2015}.

Using mask-level region proposals, \emph{instance-aware semantic segmentation} can be addressed based on the R-CNN philosophy, as in R-CNN \cite{Girshick2014}, SDS \cite{Hariharan2014}, and Hypercolumn \cite{Hariharan2015}. Sharing convolutional features among mask-level proposals is enabled by using masking layers \cite{Dai2015}. All these methods \cite{Girshick2014,Hariharan2014,Hariharan2015,Dai2015} rely on computationally expensive mask proposal methods. For example, the widely used MCG \cite{Arbelaez2014} takes 30 seconds processing an image, which becomes a bottleneck at inference time. DeepMask \cite{Pinheiro2015} is recently developed for learning segmentation candidates using convolutional networks, taking over 1 second per image. Its accuracy for instance-aware semantic segmentation is yet to be evaluated.

Category-wise \emph{semantic segmentation} is elegantly tackled by end-to-end training FCNs \cite{Long2015}. The output of an FCN consists of multiple score maps, each of which is for one category. This formulation enables per-pixel regression in a fully-convolutional form, but is not able to distinguish instances of the same category. The FCN framework has been further improved in many papers (\eg, \cite{Chen2015,Zheng2015}), but these methods also have the limitations of not being able to predict instances.

\section{Multi-task Network Cascades}

In our MNC model, the network takes an image of arbitrary size as the input, and outputs instance-aware semantic segmentation results. The cascade has three stages: proposing box-level instances, regressing mask-level instances, and categorizing each instance.
These three stages are designed to share convolutional features (\eg, the 13 convolutional layers in VGG-16 \cite{Simonyan2015}).
Each stage involves a loss term, but a later stage's loss relies on the output of an earlier stage, so the three loss terms are not independent. We train the entire network cascade end-to-end with a unified loss function.
Fig.~\ref{fig:overview} illustrates our cascade model.

In this section we describe the definition for each stage. In the next section we introduce an end-to-end training algorithm to address the causal dependency.

\subsection{Regressing Box-level Instances}

In the first stage, the network proposes object instances in the form of bounding boxes. These bounding boxes are class-agnostic, and are predicted with an objectness score.

The network structure and loss function of this stage follow the work of Region Proposal Networks (RPNs) \cite{Ren2015}, which we briefly describe as follows for completeness. An RPN predicts bounding box locations and objectness scores in a fully-convolutional form. On top of the shared features, a 3$\times$3 convolutional layer is used for reducing dimensions, followed by two sibling 1$\times$1 convolutional layers for regressing box locations and classifying object/non-object. The box regression is with reference to a series of pre-defined boxes (called ``anchors'' \cite{Ren2015}) at each location.

We use the RPN loss function given in \cite{Ren2015}. This loss function serves as the loss term $L_1$ of our stage 1.
It has a form of:
\begin{equation}\label{eq:L1}
L_1 = L_1(B(\Theta)).
\end{equation}
Here $\Theta$ represents all network parameters to be optimized. $B$ is the network output of this stage, representing a list of boxes: $B=\{B_i\}$ and $B_i=\{x_i, y_i, w_i, h_i, p_i\}$, where $B_i$ is a box indexed by $i$. The box $B_i$ is centered at $(x_i, y_i)$ with width $w_i$ and height $h_i$, and $p_i$ is the objectness probability.
The notations in Eqn.(\ref{eq:L1}) indicate that the box predictions are functions of the network parameters $\Theta$.

\subsection{Regressing Mask-level Instances}

The second stage takes the shared convolutional features and stage-1 boxes as input. It outputs a pixel-level segmentation mask for each box proposal. In this stage, a mask-level instance is still class-agnostic.

Given a box predicted by stage 1, we extract a feature of this box by Region-of-Interest (RoI) pooling \cite{He2014,Girshick2015}.
The purpose of RoI pooling is for producing a fixed-size feature from an arbitrary box, which is set as 14$\times$14 at this stage.
We append two extra fully-connected (fc) layers to this feature for each box. The first fc layer (with ReLU) reduces the dimension to 256, followed by the second fc layer that regresses a pixel-wise mask. This mask, of a pre-defined spatial resolution of $m\times m$ (we use $m=28$), is parameterized by an $m^2$-dimensional vector. The second fc layer has $m^2$ outputs, each performing binary logistic regression to the ground truth mask.

With these definitions, the loss term $L_2$ of stage 2 for regressing masks exhibits the following form:
\begin{equation}\label{eq:L2}
L_2 = L_2(M(\Theta) ~|~B(\Theta)).
\end{equation}
Here $M$ is the network outputs of this stage, representing a list of masks: $M=\{M_i\}$ and $M_i$ is an $m^2$-dimensional logistic regression output (via sigmoid) taking continuous values in $[0, 1]$. Eqn.(\ref{eq:L2}) indicates that the mask regression loss $L_2$ is dependent on $M$ but also on $B$.

As a related method, DeepMask \cite{Pinheiro2015} also regresses discretized masks. DeepMask applies the regression layers to dense sliding windows (fully-convolutionally), but our method only regresses masks from a few proposed boxes and so reduces computational cost.
Moreover, mask regression is only one stage in our network cascade that shares features among multiple stages, so the marginal cost of the mask regression layers is very small.

\subsection{Categorizing Instances}

The third stage takes the shared convolutional features, stage-1 boxes, and stage-2 masks as input. It outputs category scores for each instance.

Given a box predicted by stage 1, we also extract a feature by RoI pooling. This feature map is then ``masked'' by the stage-2 mask prediction, inspired by the feature masking strategy in \cite{Dai2015}. This leads to a feature focused on the foreground of the prediction mask.
The masked feature is given by element-wise product:
\begin{equation}\label{eq:masking}
\mathcal{F}^{Mask}_i(\Theta)=\mathcal{F}^{RoI}_i(\Theta)\cdot M_i(\Theta).
\end{equation}
Here $\mathcal{F}^{RoI}_i$ is the feature after RoI pooling, $M_i(\Theta)$ is a mask prediction from stage 2 (resized to the RoI resolution), and $\cdot$ represents element-wise product. The masked feature $\mathcal{F}^{Mask}_i$ is dependent on $M_i(\Theta)$.
Two 4096-d fc layers are applied on the masked feature $\mathcal{F}^{Mask}_i$. This is a mask-based pathway. Following \cite{Hariharan2014}, we also use another box-based pathway, where the RoI pooled features directly fed into two 4096-d fc layers (this pathway is not illustrated in Fig.~\ref{fig:overview}). The mask-based and box-based pathways are concatenated. On top of the concatenation, a softmax classifier of $N$+1 ways is used for predicting $N$ categories plus one background category. The box-level pathway may address the cases when the feature is mostly masked out by the mask-level pathway (\eg, on background).

The loss term $L_3$ of stage 3 exhibits the following form:
\begin{equation}\label{eq:L3}
L_3 = L_3(C(\Theta) ~|~B(\Theta), M(\Theta)).
\end{equation}
Here $C$ is the network outputs of this stage, representing a list of category predictions for all instances: $C=\{C_i\}$. This loss term is dependent on $B(\Theta)$ and $M(\Theta)$ (where $B(\Theta)$ is used for generating the RoI feature).

\section{End-to-End Training}
\label{sec:training}

We define the loss function of the entire cascade as:
\begin{equation}
\begin{aligned}
L(\Theta) = & L_1(B(\Theta)) + L_2(M(\Theta)~|~B(\Theta)) \\
			 & + L_3(C(\Theta)~|~B(\Theta), M(\Theta)),
\end{aligned}
\label{eq:exact_loss}
\end{equation}
where balance weights of 1 are implicitly used among the three terms.
$L(\Theta)$ is minimized \wrt the network parameters $\Theta$.
This loss function is \emph{unlike} traditional multi-task learning, because the loss term of a later stage depends on the output of the earlier ones. For example, based on the chain rule of backpropagation, the gradient of $L_2$ involves the gradients \wrt $B$.

The main technical challenge of applying the chain rule to Eqn.(\ref{eq:exact_loss}) lies on the spatial transform of a predicted box $B_i(\Theta)$ that determines RoI pooling. For the RoI pooling layer, its inputs are a predicted box $B_i(\Theta)$ and the convolutional feature map $\mathcal{F}(\Theta)$, both being functions of $\Theta$. In Fast R-CNN \cite{Girshick2015}, the box proposals are pre-computed and fixed, and the backpropagation of RoI pooling layer in \cite{Girshick2015} only involves $\mathcal{F}(\Theta)$. However, this is not the case in the presence of $B(\Theta)$. Gradients of both terms need to be considered in a theoretically sound end-to-end training solution.

In this section, we develop a differentiable RoI warping layer to account for the gradient \wrt predicted box positions and address the dependency on $B(\Theta)$. The dependency on $M(\Theta)$ is also tackled accordingly.

\vspace{.5em}
\noindent\textbf{Differentiable RoI Warping Layers.}
The RoI pooling layer \cite{Girshick2015,He2014} performs max pooling on a discrete grid based on a box. To derive a form that is differentiable \wrt the box position, we perform RoI pooling by a differentiable RoI \emph{warping} layer followed by standard max pooling.

The RoI warping layer crops a feature map region and warps it into a target size by interpolation.
We use $\mathcal{F}(\Theta)$ to denote the full-image convolutional feature map. Given a predicted box $B_i(\Theta)$ centered at $(x_i(\Theta), y_i(\Theta))$ with width $w_i(\Theta)$ and height $h_i(\Theta)$, an RoI warping layer interpolates the features inside the box and outputs a feature of a fixed spatial resolution. This operation can be written as linear transform on the feature map $\mathcal{F}(\Theta)$:
\begin{equation}\label{eq:warping}
\mathcal{F}^{RoI}_i(\Theta)=G(B_i(\Theta))\mathcal{F}(\Theta).
\end{equation}
Here $\mathcal{F}(\Theta)$ is reshaped as an $n$-dimensional vector, with $n=WH$ for a full-image feature map of a spatial size $W\times H$. $G$ represents the cropping and warping operations, and is an $n'$-by-$n$ matrix where $n'=W'H'$ corresponds to the pre-defined RoI warping output resolution $W'\times H'$.
$\mathcal{F}^{RoI}_i(\Theta)$ is an $n'$-dimensional vector representing the RoI warping output. We note that these operations are performed for each channel independently.

The computation in Eqn.(\ref{eq:warping}) has this form:
\begin{equation}\label{eq:warping1}
\mathcal{F}^{RoI}_i{_{(u', v')}}=\sum^{W\times H }_{(u,v)}G(u, v; u', v' | B_i)\mathcal{F}_{(u, v)},
\end{equation}
where the notations $\Theta$ in Eqn.(\ref{eq:warping}) are omitted for simplifying presentation. Here $(u', v')$ represent a spatial position in the target $W'\times H'$ feature map, and $(u, v)$ run over the full-image feature map $\mathcal{F}$.

The function $G(u, v; u', v' | B_i)$ represents transforming a proposed box $B_i$ from a size of $[x_i-w_i/2, x_i+w_i/2)\times[y_i-h_i/2, y_i+h_i/2)$ into another size of $[-W'/2, W'/2)\times[-H'/2, H'/2)$. Using bilinear interpolation, $G$ is separable: $G(u, v; u', v' | B_i)=g(u, u' | x_i, w_i)g(v, v' | y_i, h_i)$ where:
\vspace{-.5em}
\begin{equation}
g(u, u' | x_i, w_i) = \kappa(x_i+\frac{u'}{W'}w_i-u),
\label{eq:warping2}
\end{equation}
where $\kappa(\cdot)=max(0, 1-|\cdot|)$ is the bilinear interpolation function, and $x_i+\frac{u'}{W'}w_i$ maps the position of $u'\in[-W'/2, W'/2)$ to the full-image feature map domain.
$g(v, v' | y_i, h_i)$ is defined similarly. We note that because $\kappa$ is non-zero in a small interval, the actual computation of Eqn.(\ref{eq:warping1}) involves a very few terms.

According to the chain rule, for backpropagation involving Eqn.(\ref{eq:warping}) we need to compute:
\begin{equation}
\frac{\partial{L_2}}{\partial{B_i}} = 
\frac{\partial{L_2}}{\partial{\mathcal{F}^{RoI}_i}} \frac{\partial{G}}{\partial{B_i}}\mathcal{F}
\label{eq:RoIwarping}
\end{equation}
where we use ${\partial{B_i}}$ to denote ${\partial{x_i}}$, ${\partial{y_i}}$, ${\partial{w_i}}$, and ${\partial{h_i}}$ for simplicity. The term $\frac{\partial{G}}{\partial{B_i}}$ in Eqn.(\ref{eq:RoIwarping}) can be derived from Eqn.(\ref{eq:warping2}). As such, the RoI warping layer can be trained with any preceding/succeding layers. If the boxes are constant (\eg, given by Selective Search \cite{Uijlings2013}), Eqn.(\ref{eq:RoIwarping}) is not needed, which becomes the case of the existing RoI pooling in \cite{Girshick2015}.

After the differentiable RoI warping layer, we append a max pooling layer to perform the RoI max pooling behavior. We expect the RoI warping layer to produce a sufficiently fine resolution, which is set as $W'\times H'=28\times28$ in this paper. A max pooling layer is then applied to produce a lower-resolution output, \eg, 7$\times$7 for VGG-16.

The RoI warping layer shares similar motivations with the recent work of Spatial Transformer Networks \cite{Jaderberg2015}. In \cite{Jaderberg2015}, a spatial transformation of the entire image is learned, which is done by feature interpolation that is differentiable \wrt the transformation parameters. The networks in \cite{Jaderberg2015} are used for image classification. Our RoI warping layer is also driven by the differentiable property of interpolating features.
But the RoI warping layer is applied to multiple proposed boxes that are of interest, instead of the entire image. The RoI warping layer has a pre-defined output size and arbitrary input sizes, in contrast to \cite{Jaderberg2015}.

\vspace{.5em}
\noindent\textbf{Masking Layers.}
We also compute the gradients involved in $L_3(C(\Theta)~|~B(\Theta), M(\Theta))$, where the dependency on $B(\Theta)$ and $M(\Theta)$ is determined by Eqn.(\ref{eq:masking}).
With the differentiable RoI warping module ($\mathcal{F}^{RoI}_i$), the operations in Eqn.(\ref{eq:masking}) can be simply implemented by an element-wise product module.

\vspace{1em}
In summary, given the differentiable RoI warping module, we have all the necessary components for backpropagation (other components are either standard, or trivial to implement). We train the model by stochastic gradient descent (SGD), implemented in the Caffe library \cite{Jia2014}.

\section{Cascades with More Stages}
\label{sec:morestages}

Next we extend the cascade model to more stages within the above MNC framework.

In Fast R-CNN \cite{Girshick2015}, the ($N$+1)-way classifier is trained jointly with class-wise bounding box regression. Inspired by this practice, on stage 3, we add a 4($N$+1)-d fc layer for regression class-wise bounding boxes \cite{Girshick2015}, which is a sibling layer with the classifier layer. The entire 3-stage network cascade is trained as in Sec.~\ref{sec:training}.

\begin{figure}
\begin{center}
\includegraphics[width=0.85\linewidth]{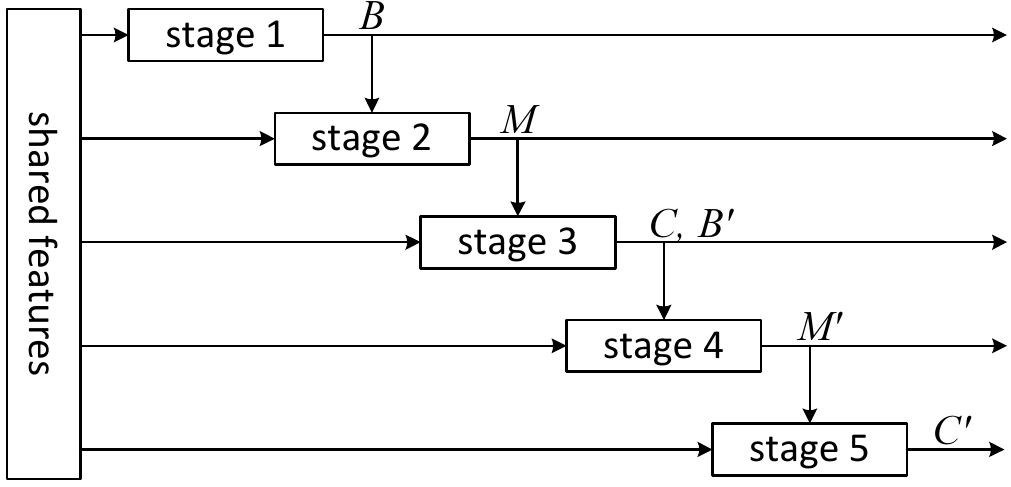}
\end{center}
\vspace{-1em}
\caption{A 5-stage cascade. On stage 3, bounding boxes updated by the box regression layer are used as the input to stage 4.}
\label{fig:5stage}
\end{figure}

The inference step with box regression, however, is not as straightforward as in object detection, because our ultimate outputs are masks instead of boxes. So during inference, we first run the entire 3-stage network and obtain the regressed boxes on stage 3. These boxes are then considered as new proposals\footnote{To avoid multiplying the number of proposals by the number of categories, for each box we only use the highest scored category's bounding box regressor.}.
Stages 2 and 3 are performed for the second time on these proposals.
This is in fact \emph{5-stage inference}. Its inference-time structure is illustrated in Fig.~\ref{fig:5stage}. The new stages 4 and 5 share the same structures as stages 2 and 3, except that they use the regressed boxes from stage 3 as the new proposals. This inference process can be iterated, but we have observed negligible gains.

Given the above \emph{5-stage cascade} structure (Fig.~\ref{fig:5stage}), it is easy to adopt our algorithm in Sec.~\ref{sec:training} to train this cascade end-to-end by backpropagation. Training the model in this way makes the training-time structure consistent with the inference-time structure, which improves accuracy as will be shown by experiments.
It is possible to train a cascade with even more stages in this way. But due to concerns on fast inference, we only present MNCs with up to 5 stages.

\setlength{\tabcolsep}{7pt}
\renewcommand{\arraystretch}{1.05}
\begin{table*}[t]
\begin{center}
\small
\begin{tabular}{c|cccc|cccc}
 & \multicolumn{4}{c|}{ZF net} & \multicolumn{4}{c}{VGG-16 net} \\
\hline
training strategies & (a) & (b) & (c) & (d) & (a) & (b) & (c) & (d)\\
\hline
shared features? & & \checkmark & \checkmark & \checkmark & & \checkmark & \checkmark & \checkmark\\
end-to-end training? & & & \checkmark & \checkmark & & & \checkmark & \checkmark \\
training 5-stage cascades? & & & & \checkmark & & & & \checkmark \\
\hline
mAP$^r$@0.5 (\%) & 51.8 & 52.2 & 53.5 & \textbf{54.0} & 60.2 & 60.5 & 62.6 & \textbf{63.5} \\
\end{tabular}
\end{center}
\vspace{-.5em}
\caption{Ablation experiments on PASCAL VOC 2012 validation. For (a), (b), and (c), the cascade structures for training have 3 stages. The inference process (5-stage, see~\ref{sec:morestages}) is the same for all cases; the models are only different in the training methods. The pre-trained models are ZF net \cite{Zeiler2014} (left) and VGG-16 net \cite{Simonyan2015} (right).}
\label{tab:ablation_result}
\end{table*}

\section{Implementation Details}

\noindent\emph{Non-maximum suppression.}
On stage 1, the network produces $\sim$$10^4$ regressed boxes. For generating the proposals for stage 2, we use non-maximum suppression (NMS) to reduce redundant candidates. The threshold of the Intersection-over-Union (IoU) ratio for this NMS is 0.7 as in \cite{Ren2015}. After that, the top-ranked 300 boxes \cite{Ren2015} will be used for stage 2. During training, the forward/backward propagated signals of stages 2 and 3 only go through the ``pathways'' determined by these 300 boxes. NMS is similar to max pooling, maxout \cite{Goodfellow2013}, or other local competing layers \cite{Srivastava2013}, which are implemented as routers of forward/backward pathways.
During inference, we use the same NMS strategy to produce 300 proposals for stage 2.

\vspace{.5em}
\noindent\emph{Positive/negative samples.} (i) On stage 1, their definitions follow \cite{Ren2015}. (ii) On stage 2, for each proposed box we find its highest overlapping ground truth mask. If the overlapping ratio (IoU) is greater than 0.5, this proposed box is considered as positive and contributes to the mask regression loss; otherwise is ignored in the regression loss. The mask regression target is the intersection between the proposed box and the ground truth mask, resized to $m\times m$ pixels. (iii) On stage 3, we consider two sets of positive/negative samples. In the first set, the positive samples are the instances that overlap with ground truth boxes by \emph{box-level} IoU $\geq 0.5$ (the negative samples are the rest). In the second set, the positive samples are the instances that overlap with ground truth instances by box-level IoU $\geq 0.5$ \emph{and} mask-level IoU $\geq 0.5$.
The loss function of stage 3 involves two ($N$+1)-way classifiers, one for classifying mask-level instances and the other for classifying box-level instances (whose scores are not used for inference). 
The reason for considering both box-level and mask-level IoU is that when the proposed box is not a real instance (\eg, on the background or poorly overlapping with ground truth), the regressed mask might be less reliable and thus the box-level IoU is more confident.

\vspace{.5em}
\noindent\emph{Hyper-parameters for training.}
We use the ImageNet pre-trained models (\eg, VGG-16 \cite{Simonyan2015}) to initialize the shared convolutional layers and the corresponding 4096-d fc layers. The extra layers are initialized randomly as in \cite{He2015}. We adopt an \emph{image-centric} training framework \cite{Girshick2015}: the shared convolutional layers are computed on the entire image, while the RoIs are randomly sampled for computing loss functions. In our system, each mini-batch involves 1 image, 256 sampled anchors for stage 1 as in \cite{Ren2015}\footnote{Though we sample 256 anchors on stage 1 for computing the loss function, the network of stage 1 is still computed fully-convolutionally on the entire image and produces all proposals that are used by later stages.}, and 64 sampled RoIs for stages 2 and 3. We train the model using a learning rate of 0.001 for 32k iterations, and 0.0001 for the next 8k. We train the model in 8 GPUs, each GPU holding 1 mini-batch (so the effective mini-batch size is $\times$8). The images are resized such that the shorter side has 600 pixels \cite{Girshick2015}. We do not adopt multi-scale training/testing \cite{He2014,Girshick2015}, as it provides no good trade-off on speed \vs accuracy \cite{Girshick2015}.

\vspace{.5em}
\noindent\emph{Inference.}
We use 5-stage inference for both 3-stage and 5-stage trained structures. The inference process gives us a list of 600 instances with masks and category scores (300 from the stage 3 outputs, and 300 from the stage 5 outputs).
We post-process this list to reduce similar predictions. We first apply NMS (using box-level IoU 0.3 \cite{Girshick2014}) on the list of 600 instances based on their category scores. After that, for each not-suppressed instance, we find its ``similar'' instances which are defined as the suppressed instances that overlap with it by IoU $\geq$ 0.5. The prediction masks of the not-suppressed instance and its similar instances are merged together by weighted averaging, pixel-by-pixel, using the classification scores as their averaging weights. This ``mask voting'' scheme is inspired by the box voting in \cite{Gidaris2015}.
The averaged masks, taking continuous values in $[0, 1]$, are binarized to form the final output masks. The averaging step improves accuracy by $\sim$1\% over the NMS outcome.
This post-processing is performed for each category independently.

\section{Experiments}

\subsection{Experiments on PASCAL VOC 2012}

We follow the protocols used in recent papers \cite{Hariharan2014,Dai2015,Hariharan2015} for evaluating instance-aware semantic segmentation. The models are trained on the PASCAL VOC 2012 training set, and evaluated on the validation set. We use the segmentation annotations in \cite{Hariharan2011} for training and evaluation, following \cite{Hariharan2014,Dai2015,Hariharan2015}.
We evaluate the mean Average Precision, which is referred to as mean AP$^r$ \cite{Hariharan2014} or simply mAP$^r$. We evaluate mAP$^r$ using IoU thresholds at 0.5 and 0.7.

\vspace{.5em}
\noindent\textbf{Ablation Experiments on Training Strategies.} Table \ref{tab:ablation_result} compares the results of different training strategies for MNCs. We remark that in this table all results are obtained via 5-stage inference, so the differences are contributed by the training strategies. We show results using ZF net \cite{Zeiler2014} that has 5 convolutional layers and 3 fc layers, and VGG-16 net \cite{Simonyan2015} that has 13 convolutional layers and 3 fc layers.

As a simple baseline (Table \ref{tab:ablation_result}, a), we train the three stages step-by-step \emph{without} sharing their features. Three separate networks are trained, and a network of a later stage takes the outputs from the trained networks of the earlier stages. The three separate networks are all initialized by the ImageNet-pre-trained model. This baseline has an mAP$^r$ of 60.2\% using VGG-16. We note that \emph{this baseline result is competitive} (see also Table~\ref{tab:voc_result}), suggesting that decomposing the task into three sub-tasks is an effective solution.

To achieve feature sharing, one may follow the step-by-step training in \cite{Ren2015}. Given the above model (a), the shared convolutional layers are kept unchanged by using the last stage's weights, and the three separate networks are trained step-by-step again with the shared layers not tuned, following \cite{Ren2015}. Doing so leads to an mAP$^r$ of 60.5\%, just on par with the baseline that does not share features. This suggests that sharing features does not directly improve accuracy.

Next we experiment with the single-step, end-to-end training algorithm developed in Sec.~\ref{sec:training}. Table~\ref{tab:ablation_result} (c) shows the result of end-to-end training a 3-stage cascade. The mAP$^r$ is increased to 62.6\%. We note that in Table ~\ref{tab:ablation_result} (a), (b), and (c), the models have the same structure for training. So the improvement of (c) is contributed by end-to-end training this cascade structure.
This improvement is similar to other gains observed in many practices of multi-task learning \cite{Caruana1997}.
By developing training algorithm as in Sec.~\ref{sec:training}, we are able to train the network by backpropagation in a theoretically sound way. The features are naturally shared by optimizing a unified loss function, and the benefits of multi-task learning are witnessed.

Table~\ref{tab:ablation_result} (d) shows the result of end-to-end training a 5-stage cascade. The mAP$^r$ is further improved to 63.5\%. We note that all results in Table ~\ref{tab:ablation_result} are based on the same 5-stage inference strategy. So the accuracy gap between (d) and (c) is contributed by training a 5-stage structure that is \emph{consistent} with its inference-time usage.

The series of comparisons are also observed when using the ZF net as the pre-trained model (Table~\ref{tab:ablation_result}, left), showing the generality of our findings.

\setlength{\tabcolsep}{2pt}
\renewcommand{\arraystretch}{1.05}
\begin{table}[t]
\begin{center}
\small
\begin{tabular}{l|c|c|c}
method &\footnotesize mAP$^{r}$@0.5 (\%) &\footnotesize mAP$^{r}$@0.7 (\%) &\footnotesize time/img (s) \\
\hline
O$^2$P \cite{Carreira2012} & 25.2 & - & - \\
SDS (AlexNet) \cite{Hariharan2014} & 49.7 & 25.3 & 48 \\
Hypercolumn \cite{Hariharan2015} & 60.0 & 40.4 & $>$80 \\
CFM \cite{Dai2015} & 60.7 & 39.6 & 32 \\
\hline
MNC [ours] & \textbf{63.5} & \textbf{41.5} & \textbf{0.36} \\
\end{tabular}	
\end{center}
\vspace{-.5em}
\caption{Comparisons of instance-aware semantic segmentation on the PASCAL VOC 2012 validation set. The testing time per image (\emph{including all steps}) is evaluated in a single Nvidia K40 GPU, except that the MCG \cite{Arbelaez2014} proposal time is evaluated on a CPU. MCG is used by \cite{Hariharan2014,Hariharan2015,Dai2015} and its running time is about 30s. The running time of \cite{Hariharan2015} is our estimation based on the description from the paper.
The pre-trained model is VGG-16 for \cite{Hariharan2015,Dai2015} and ours. O$^2$P is not based on deep CNNs, and its result is reported by \cite{Hariharan2014}.}
\label{tab:voc_result}
\end{table}

\setlength{\tabcolsep}{6pt}
\renewcommand{\arraystretch}{1.05}
\begin{table}[t]
\begin{center}
\small
\begin{tabular}{cccccc|c}
conv & stage 2 & stage 3 & stage 4 & stage 5 & others & total \\
\hline
0.15 & 0.01 & 0.08 & 0.01 & 0.08 & 0.03 & 0.36 \\
\end{tabular}	
\end{center}
\vspace{-1em}
\caption{Detailed testing time (seconds) per image of our method using 5-stage inference. The model is VGG-16. ``Others'' include post-processing and communications among stages.}
\label{tab:time}
\end{table}

\vspace{.5em}
\noindent\textbf{Comparisons with State-of-the-art Methods.} In Table~\ref{tab:voc_result} we compare with SDS \cite{Hariharan2014}, Hypercolumn \cite{Hariharan2015}, and CFM \cite{Dai2015}, which are existing CNN-based semantic segmentation methods that are able to identify instances. These papers reported their mAP$^r$ under the same protocol used by our experiments.
Our MNC has $\sim$3\% higher mAP$^r$@0.5 than previous best results. Our method also has higher mAP$^r$@0.7 than previous methods.

\begin{figure*}
\begin{center}
\includegraphics[width=0.72\linewidth]{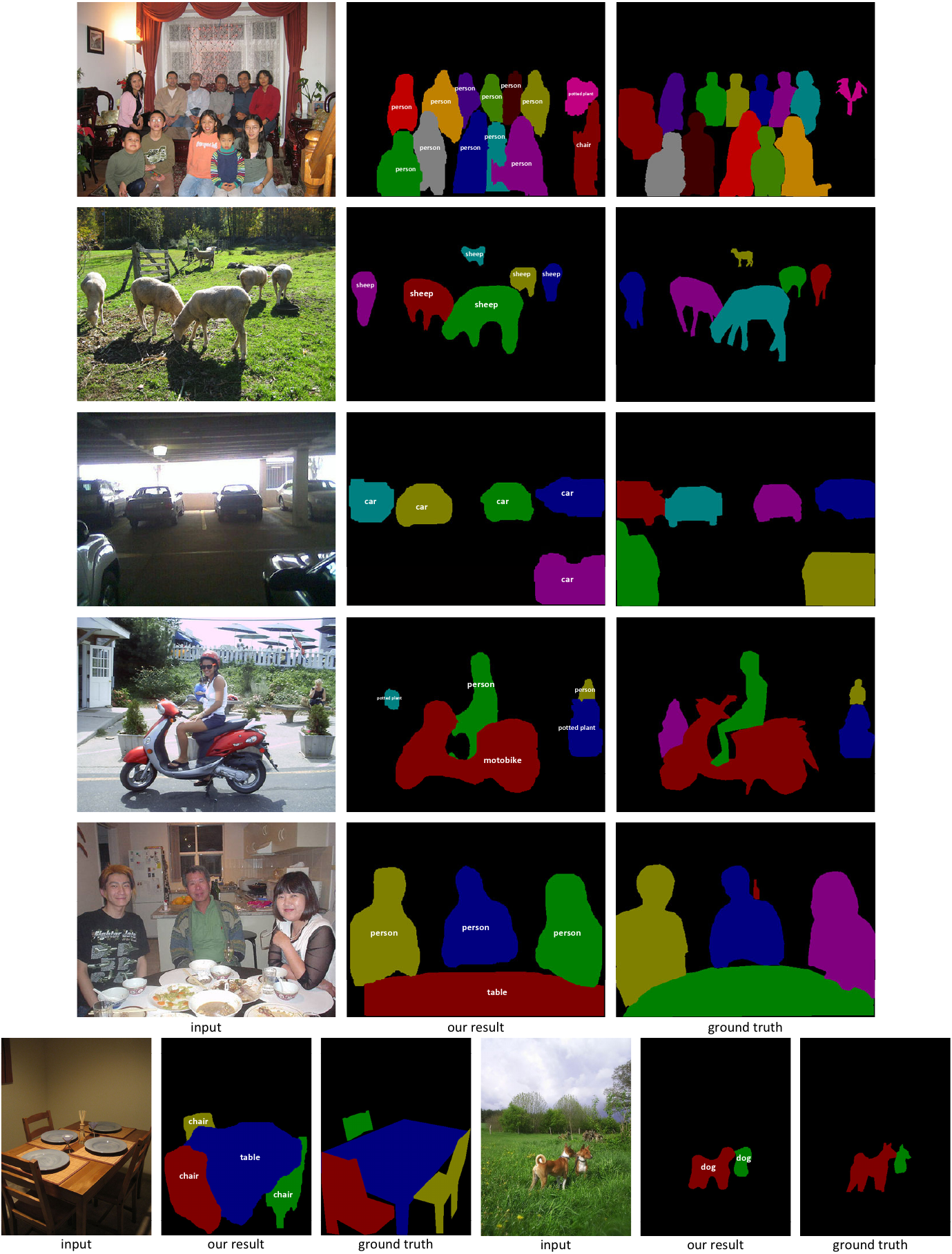}
\end{center}
\vspace{-1em}
\caption{Our instance-aware semantic segmentation results on the PASCAL VOC 2012 validation set. One color denotes one instance.}
\label{fig:voc_results}
\vspace{-1em}
\end{figure*}

Fig~\ref{fig:voc_results} shows some examples of our results on the validation set. Our method can handle challenging cases where multiple instances of the same category are spatially connected to each other (\eg, Fig~\ref{fig:voc_results}, first row).

\vspace{.5em}
\noindent\textbf{Running Time.}
Our method has an inference-time speed of \textbf{360ms} per image (Table~\ref{tab:voc_result}), evaluated on an Nvidia K40 GPU. Table~\ref{tab:time} shows the details.
Our method does not require any external region proposal method, whereas the region proposal step in SDS, Hypercolumn, and CFM costs 30s using MCG. Furthermore, our method uses the shared convolutional features for the three sub-tasks and avoids redundant computation. Our system is about two orders of magnitude faster than previous systems.

\vspace{.5em}
\noindent\textbf{Object Detection Evaluations.} We are also interested in the box-level object detection performance (mAP$^b$), so that we can compare with more systems that are designed for object detection.
We train our model on the PASCAL VOC 2012 \emph{trainval} set, and evaluate on the PASCAL VOC 2012 \emph{test} set for object detection.
Given mask-level instances generated by our model, we simply assign a tight bounding box to each instance.
Table~\ref{tab:det_result} shows that our result (70.9\%) compares favorably to the recent Fast/Faster R-CNN systems \cite{Girshick2015,Ren2015}. We note that our result is obtained with \emph{fewer training images} (without the 2007 set), but with mask-level annotations. This experiment shows the effectiveness of our algorithm for detecting both box- and mask-level instances.

The above detection result is solely based on the mask-level outputs. But our method also has box-level outputs from the box regression layers in stage 3/5. Using these box layers' outputs (box coordinates and scores) in place of the mask-level outputs, we obtain an mAP$^b$ of 73.5\% (Table~\ref{tab:det_result}).
Finally, we train the MNC model on the union set of 2007 trainval+test and 2012 trainval. As the 2007 set has no mask-level annotation, when a sample image from the 2007 set is used, its mask regression loss is ignored (but the mask is generated for the later stages) and its mask-level IoU measure for determining positive/negative samples is ignored. These samples can still impact the box proposal stage and the categorizing stage. Under this setting, we obtain an mAP$^b$ of \textbf{75.9\%} (Table~\ref{tab:det_result}), substantially better than Fast/Faster R-CNN \cite{Girshick2015,Ren2015}.

\setlength{\tabcolsep}{8pt}
\renewcommand{\arraystretch}{1.05}
\begin{table}[t]
\begin{center}
\small
\begin{tabular}{l|l|c}
system & training data & mAP$^{b}$ (\%) \\
\hline
R-CNN \cite{Girshick2014} & VOC 12 & 62.4 \\
Fast R-CNN \cite{Girshick2015} & VOC 12 &  65.7 \\
Fast R-CNN \cite{Girshick2015} & VOC 07++12 &  68.4 \\
Faster R-CNN \cite{Ren2015} & VOC 12 & 67.0 \\
Faster R-CNN \cite{Ren2015} & VOC 07++12 & 70.4 \\
\hline
MNC [ours] & VOC 12 & 70.9 \\
MNC$_{box}$ [ours] & VOC 12 & 73.5 \\
MNC$_{box}$ [ours]$^\dag$ & VOC 07++12 & \textbf{75.9} \\
\end{tabular}
\end{center}
\vspace{-.5em}
\caption{Evaluation of (box-level) object detection mAP on the PASCAL VOC 2012 test set. ``12'' denotes VOC 2012 trainval, and ``07++12'' denotes VOC 2007 trainval+test and 2012 trainval. The pre-trained model is VGG-16 for all methods.$^\dag$: \url{http://host.robots.ox.ac.uk:8080/anonymous/NUWDYX.html}}
\label{tab:det_result}
\end{table}

\setlength{\tabcolsep}{8pt}
\renewcommand{\arraystretch}{1.05}
\begin{table}[t]
\begin{center}
\small
\begin{tabular}{c|c|c}
network & \footnotesize mAP@[.5:.95] (\%)  & \footnotesize ~~~mAP@.5 (\%)~~~\\
\hline
VGG-16 \cite{Simonyan2015} & 19.5 & 39.7 \\
ResNet-101 \cite{He2015a} & \textbf{24.6} & \textbf{44.3} \\
\end{tabular}	
\end{center}
\vspace{-1em}
\caption{Our baseline segmentation result (\%) on the MS COCO \emph{test-dev} set. The training set is the \emph{trainval} set.}
\label{tab:coco_result}
\end{table}

\begin{figure*}[t]
\begin{center}
\includegraphics[width=0.95\linewidth]{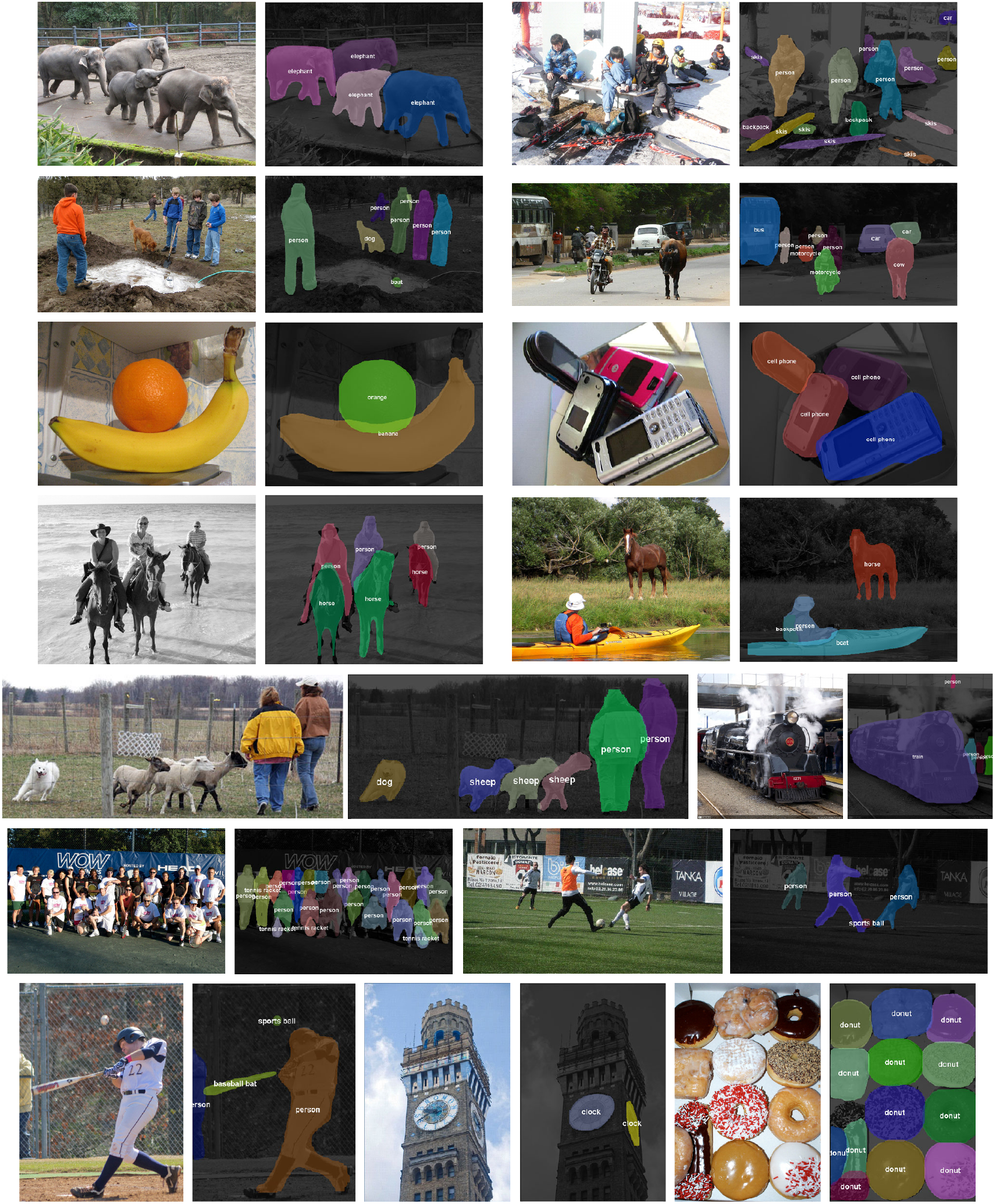}
\end{center}
\vspace{-1em}
\caption{Our instance-aware semantic segmentation results on the MS COCO test-dev set using ResNet-101 \cite{He2015a}.}
\label{fig:coco_results}
\vspace{-1em}
\end{figure*}

\subsection{Experiments on MS COCO Segmentation}

We further evaluate on the MS COCO dataset \cite{Lin2014}.
This dataset consists of 80 object categories for instance-aware semantic segmentation. Following the COCO guidelines, we use the 80k+40k \emph{trainval} images to train, and report the results on the \emph{test-dev} set. We evaluate the standard COCO metric (mAP$^r$@IoU=[0.5:0.95]) and also the PASCAL metrics (mAP$^r$@IoU=0.5). Table \ref{tab:coco_result} shows our method using VGG-16 has a result of 19.5\%/39.7\%.

\emph{The end-to-end training behavior and the independence of external models make our method easily enjoy gains from deeper representations.}
By replacing VGG-16 with an extremely deep 101-layer network (ResNet-101) \cite{He2015a}, we achieve 24.6\%/44.3\% on the MS COCO test-dev set (Table \ref{tab:coco_result}).
It is noteworthy that ResNet-101 leads to a relative improvement of 26\% (on mAP$^r$@[.5:.95]) over VGG-16, which is consistent to the relative improvement of COCO object detection in \cite{He2015a}.
This baseline result is close to the 2nd-place winner's ensemble result (25.1\%/45.8\% by FAIRCNN).
On our baseline result, we further adopt global context modeling and multi-scale testing as in \cite{He2015a}, and ensembling. Our final result on the test-challenge set is 28.2\%/51.5\%, which \emph{won the 1st place in the COCO segmentation track}\footnote{\fontsize{7pt}{1em}\selectfont\url{http://mscoco.org/dataset/\#detections-challenge2015}} of ILSVRC \& COCO 2015 competitions. Fig.~\ref{fig:coco_results} shows some examples.

\section{Conclusion}

\vspace{-.5em}

We have presented Multi-task Network Cascades for fast and accurate instance segmentation. We believe that the idea of exploiting network cascades in a multi-task learning framework is general. This idea, if further developed, may be useful for other recognition tasks.

Our method is designed with fast inference in mind, and is orthogonal to some other successful strategies developed previously for semantic segmentation. For example, one may consider exploiting a CRF \cite{Chen2015} to refine the boundaries of the instance masks. This is beyond the scope of this paper and will be investigated in the future.

{\small
\bibliographystyle{ieee}
\bibliography{ins_seg_arxiv_v1_release}
}

\end{document}